# A Framework for Pre-processing of Social Media Feeds based on Integrated Local Knowledge Base

Taiwo Kolajo [a, b*], Olawande Daramola [c], Ayodele Adebiyi [a, d], and Seth Aaditeshwar [e]

*[a]Department of Computer and Information Sciences, Covenant University, Ota, Nigeria.
taiwo.kolajo@stu.cu.edu.ng, ayo.adebiyi@covenantuniversity.edu.ng
https://orcid.org/0000-0001-6780-2945*

*[b]Department of Computer Science, Federal University Lokoja, Kogi State, Nigeria
taiwo.kolajo@fulokoja.edu.ng
https://orcid.org/0000-0001-6780-2945*

*[c]Department of Information Technology, Cape Peninsula University of Technology, Cape Town, South Africa
Email: daramolaj@cput.ac.za
https://orcid.org/0000-0001-6340-078X*

*[d]Department of Computer Science, Landmark University, Omu-Aran, Kwara State, Nigeria
Email: ayo.adebiyi@lmu.edu.ng
https://orcid.org/0000-0002-3114-6315*

*[e]Department of Computer Science & Engineering, Indian Institute of Technology Delhi, New Delhi, India.
Email: aseth@cse.iitd.ac.in
https://orcid.org/0000-0001-9012-5656*

## Abstract

Most of the previous studies on the semantic analysis of social media feeds have not considered the issue of ambiguity that is associated with slangs, abbreviations, and acronyms that are embedded in social media posts. These noisy terms have implicit meanings and form part of the rich semantic context that must be analysed to gain complete insights from social media feeds. This paper proposes an improved framework for pre-processing of social media feeds for better performance. To do this, the use of an integrated knowledge base (*ikb*) which comprises a local knowledge source (Naijalingo), urban dictionary and internet slang was combined with the adapted Lesk algorithm to facilitate semantic analysis of social media feeds. Experimental results showed that the proposed approach performed better than existing methods when it was tested on three machine learning models, which are support vector machines, multilayer perceptron, and convolutional neural networks. The framework had an accuracy of 94.07% on a standardized dataset, and 99.78% on localised dataset when used to extract sentiments from tweets. The improved performance on the localised dataset reveals the advantage of integrating the use of local knowledge sources into the process of analysing social media feeds particularly in interpreting slangs/acronyms/abbreviations that have contextually rooted meanings.

## 1. Introduction

Social media has been identified as an important open channel of sourcing information [1]. People tend to communicate information more regularly and faster via social media due to its real-time and lightweight nature than through short message service (SMS) from the cell telephone [2,3].

Unlike painstakingly created news and different literary contents of the web, social media posts present numerous new difficulties for analytics algorithms due to their noisy, full-size scale, and social nature. Most of the contents on social media streams contain casual usage of terms, incomplete statements, noisy data, slangs, statements with spelling and grammatical mistakes, irregular sentences, abbreviated phrases/words, and improper sentence structure [4-6]. These challenges make it vital to seek improvement of the performance of existing pre-processing solutions for social media streams [7-9].

The nature of social media streams makes it difficult to understand them as stand-alone statements without reference to information from external sources [10]. Grammatical structures which are easily understood by humans cannot be easily recognized by machines and more particularly in the case of social media feeds [11]. Although efforts were made to address this issue, current knowledge-based strategies for analysing social media feeds have limitations. Most methods can only cope with shallower keyword or topic extraction, the problem with this is that such markers or

keywords of interest may not exist in some informational tweets [12]. Moreover, the majority of knowledge-based approaches for analysing slang/abbreviation/acronym do not take care of ambiguity issues that evolve in the utilisation of these noisy phrases [13]. Just like ambiguity exist with the use of ordinary language, there is also ambiguity in the utilisation of slangs/abbreviations/acronyms that needs to be rightly interpreted to enhance the results of social media analysis. For instance, the word "pale" is ambiguous, when used on social media. In the Urban Dictionary[1], it means "having a pale skin", "to be a platonic soulmate", and "a really bad hangover" whereas within the Nigerian social media circles, the word "pale" is a pidgin word (pronounced as "per le") that is synonymous in meaning to biological father. Thus, the ambiguity problem does not only apply to regular Standard English words but also the content of social media feeds. Most of the existing approaches have not addressed this problem of ambiguity in social media feeds. As a result of this, knowledge-based approaches must have the capability to semantically analyse the noisy and syntactically irregular language of social media [14,15].

This paper extends a previous paper [16] where an abridged report was presented. This study is based on the use of more elaborate data to derive a robust conclusion on the importance of resolving ambiguity in the usage of slangs/acronyms/abbreviations in social media feeds. Specifically, the following extensions were undertaken relative to [16]:

1) An open and standardised Twitter dataset, and a localised dataset of tweets of Nigerian origin (*Naija-tweets*) were used for the evaluation of the proposed framework instead of just the localised dataset that was used in [16].

2) A bigger and improved version of localised dataset (*Naija-tweets*) was used. The new *Naija-tweets* dataset contains additional 2,920 tweets, and is a balanced dataset containing an even distribution of tweets with positive and negative sentiments compared to the one in [16] that is an imbalanced dataset. The improvement of the size and content of the localised dataset is to enable a more accurate basis for the evaluation of the proposed framework.

---

[1] https://www.urbandictionary.com/

3) A proposed framework has been enhanced with the addition of an automatic spell-checking module to improve the quality of output of the framework.

4) Further elucidation of the adapted Lesk algorithm that was used for disambiguation of social media slangs/abbreviations/acronyms is presented, with an illustration example to foster a better understanding of the disambiguation process.

The adopted approach incorporates the use of local knowledge-base to analyse the contextual meanings that are hidden in the slangs, abbreviations, and acronyms that are contained in social media streams. This is because a better understanding and interpretation of the rich semantics and contextual meanings that are hidden in social media feeds will offer critical information for algorithms or techniques that builds on social media data for useful operations like event detection or sentiments analysis [17]. The objectives of this paper are to:

1) Investigate the effect of using integrated knowledge sources for the pre-processing of social media feeds and its performance on computational algorithms that rely on such for their results.

2) To determine whether resolving ambiguity in the usage of slangs/acronyms/abbreviations will lead to improved accuracy of social media analytics.

3) To investigate the impact of local knowledge source in analysing a relevant localised dataset.

As a contribution, this paper proposes an improved approach to pre-processing of social media streams by (1) capturing the full meaning, and literal essence of slangs, abbreviations and acronyms that are used in social media feeds; and (2) introducing the use of localised knowledge sources to decipher contextual knowledge in a way that ensures better and more accurate interpretation of social media streams.

The rest of this paper is structured as follows. Section 2 discusses the background and related work, Section 3 presents the methodology, while the evaluation experiment is presented in Section 4. The results of the study are presented and discussed in Section 5, while Section 6 concludes the paper with an overview of further work.

## 2.    Background and Related Work

Two of the major applications of social media data are sentiments analysis/opinion mining and event detection [18,19]. Opinion mining is used to extract the opinion of people on issues and topics of interest [14], while event detection is used to report what is happening [20]. These two major application areas of social media data have several other subcategories such as product/service improvement, brand awareness, advertising/marketing strategies, identification of trends, network structure analysis, news propagation, disaster and emergency management, political campaigns, security, press clipping, entity extraction, and more. This section presents an overview of sentiment analysis and event detection. It also includes a review of related work.

### 2.1  Sentiment analysis

Performing sentiments analysis in social media is quite challenging for natural language processing because of the intricacy and variability within the dialect articulation coupled with the availability of real-time content [21]. The first applications of sentiments analysis were on longer texts such as emails, letters, and longer web contents. Applying sentiment analysis to the microblogging fraternity is a challenging job due to the inherent noisy characteristics of social media feeds such as frequent usage of slangs, abbreviation, acronyms, incomplete statements, spelling and grammatical errors, emoticons, and hashtags amongst others [22].

Sentiment analysis has gained much attention because of the explosion of internet applications such as forums, microblogging websites, and social networks. The approaches to sentiment analysis fall into 3 categories [23], which are knowledge-based/lexical approaches [24], statistical/machine learning approaches [25], and the combination of both (hybrid) approaches [26-28]. In this paper, a combination of machine learning and knowledge-based approaches (viz. hybrid approach) to remedy ambiguity in slangs, abbreviations, acronyms to extract sentiment from tweets was employed.  Also, the use of local knowledge source has been introduced to existing knowledge sources to specifically analyse and interpret the localised slangs of Nigerian origin.

## 2.2 Event detection

Recently, event detection has received significant attention because of the prevalence of social media and its wide application in decision making and crisis management [29]. Event detection is aimed at organising a social media stream into sets of documents in such a way that each set is coherently discussing the same event. Understanding and analysing social media posts is challenging due to the presence of unrelated, noisy, and different format data [30]. Event types range from a planned event with a predefined location and time, for instance, a concert; sudden or unplanned event, (e.g., terror attack or earthquake); breaking news discussed on social media platforms; a local event that is confined to a specific geographical area (e.g. automobile accident); and entity related event (e.g. a popular singer with a new video clip) [4,31-32].

Event detection can also be categorised into specified and unspecified events. A specified event depends on known specific characteristics and information about the event which may include time, venue, type, and description. This information can be gathered from an event content material or through social media users. An unspecified event is based totally on the temporal alerts or signals of the social media streams in detecting real-world event occurrence [33].

## 2.3 Related work

This section discusses some of the previous research efforts that are associated with pre-processing/ambiguity handling in sentiment analysis and event detection. These are discussed in the following subsections.

### 2.3.1 Ambiguity handling in social media streams for sentiment analysis

The impact and effect of pre-processing which involves activities such as tokenization, removal of stop-words, lemmatization, slang analysis, and redundancy elimination on the accuracy of the result of techniques that rely on the analysis of social media data for sentiment analysis have been studied by many researchers. There is a general belief that once social media stream contents are well interpreted and represented, it leads to improvement of sentiment analysis results.

[34] presented the position of text pre-processing in sentiment analysis using a lexicon-based approach. The pre-processing stages entail stop-words removal, removal of HTML tags, URL, mentions, emoticons. A manually created dictionary that contains positive and negative words was used for abbreviation expansion. Emoticon and slang terms were not handled. It should be pointed out that analysing emoticons and slangs found in social media can greatly influence the result of sentiment analysis. Also, representing abbreviations based on words found in the dictionary without considering the context in which such abbreviations occur may not take care of ambiguity issues completely.

The impact of pre-processing methods on sentiment classification accuracy was explored by [35] with the usage of Twitter data. The pre-processing method targeted removal of URLs, mentions, hashtag, retweets. An n-gram was used to obtain binding of slang words to other co-existing words and conditional random fields to find significance and sentiment translation of such slang words. The result of the study showed an improvement in sentiment classification accuracy.

In the same vein, [36,37] investigated the position of text pre-processing and found that a suitable combination of different pre-processing tasks can improve the accuracy of social media streams classification.

[38] investigated sentiments analysis on a Twitter dataset. The pre-processing method alongside tokenization, stemming and lemmatization that was used consists of the replacement of URLs, user mention, negation, exclamation, and question marks with tokens. Letter repetition with one or two occurrences of a letter was replaced with two. The result shows that the pre-processing of Twitter data improves sentiment analysis techniques.

The pre-processing method adopted by [39,40] includes deletion of stop-words, punctuation, URLs, mentions, and stemming. [40] added the handling of slang conversion in their work although the authors did not state how the slang conversion was done. [41] studied the effect of pre-processing and discovered that expanding acronyms and changing negation enhanced classification while the elimination of stop-words, numbers or URLs did not result in enormous improvement. On the contrary, [20] evaluated classification accuracy based on pre-processing techniques and proved that getting rid of numbers, lemmatization, and replacing negation improves accuracy. From the assessment of the literature, most of the previous studies have not focused specifically on how to handle noisy terms such as slangs, abbreviations, and acronyms as well as resolving ambiguity when these noisy terms are used in social media feeds.

In addition to traditional text pre-processing, some authors also used external sources such as internet slang, dictionary, and UT-Dallas corpus and rule-based approach for improving text processing stages. [42] conducted research to normalise text presented in two languages, specifically in Roman and Indian and then tried to predict sentiment from the combination of the two languages. However, the ambiguity issue that erupted from the use of these two languages were not handled.

An approach for acronym discovery, enrichment, and disambiguation in short texts was proposed by [43]. Expansion of acronym was done using two measures, which are popularity expansion in news articles, and the use of contextual information surrounding the acronym with metadata information in the acronym dictionary data. The result shows that combining popularity expansion with context performs better than when used individually. However, the authors did not handle slang terms.

[6] proposed a rule induction framework for Twitter sentiment evaluation. In the preprocessing stage, slangs or abbreviations were filtered and taken as misspelt words. Aspell library in Python was used for treating these noisy terms and whenever the spelling checker was unable to figure out the correct meaning or word to replace a slang or an abbreviation, such a noisy term was discarded. Aspell is not sufficient to take care of these noisy terms and resolve ambiguity issues that are associated with the usage of slangs and abbreviations.

Slang sentiment dictionary was proposed by [44]. Slang words were extracted from the urban dictionary. Urban dictionary and related words were exploited to estimate sentiment polarity. This is tailored specifically to sentiment analysis. Our proposed preprocessing framework is generic in the sense that it can be used not only for sentiment analysis but also for event detection and any related application domain that rely on the analysis of social media feeds. Moreover, using only the urban dictionary for slang analysis may not properly take care of ambiguity in the usage of slangs.

Sentiment classification of tweets with the use of hybrid techniques (i.e. combination of Particle Swarm Optimisation (PSO), Decision Tree (DT), and Genetic Algorithm (GA)) was proposed by [11]. Although the author claimed that they did abbreviation expansion and correction of misspelt words, how this was done was not clearly explained in the paper. Also [45] proposed a hybrid technique to perform sentiment analysis on Twitter data that can be used for future recommendation. Naïve Bayes classifier was trained on labelled data to find the probability of the unlabeled data based on similar predefined data. However, the authors did not mention how the noisy data found in the tweets was handled.

The work of [46] focused on acronym removal by mapping an acronym with meaning from a lexicon. However, direct matching with available meaning from a lexicon without considering the context cannot adequately address ambiguity issues.

There are approaches where commonsense knowledge (CSK) have been used to obtain implicit knowledge of specific entities for natural language processing. In [47], common knowledge was acquired from open web knowledge sources such as DBpedia, ConceptNet5 and WordNet to process natural language texts. The application of the framework reported in [47] is found in [48] where concepts, properties, and relationship in the CSK known as WebChild were deployed. CSK was used to filter tweets and perform ordinance tweet mapping based on Smart City Characteristics (SCC) for sentiments analysis. To assess public opinions on local laws governing a particular urban region, SentiWordNet was used for polarity-based classification. Due to redundant information in these open web knowledge sources, inference from such sources can be extremely slow [49]. Also, this framework did not handle the noisy terms commonly found in tweets.

Cross-lingual propagation for deep sentiment analysis was presented by [50] to yield sentiment embedding vectors for numerous languages. While this is an appreciable effort in mitigating the scarcity of data in multiple languages, the paper is not focused on enriching the pre-processing stage of social media streams for a faster, understandable and interpretable social media data. Besides, the authors did not elaborate on how the noisy data commonly associated with social media data was handled.

## 2.3.2   Ambiguity handling in social media streams for event detection

Many of the previous efforts that focused on event detection have attended to aspects of natural language pre-processing such as tokenization, stop-word removal and stemming but not much attention has been given to noisy and abnormal phrases consisting of slangs, abbreviations, and acronyms that are rampant in social media streams. [51] presented a multimodal classification of events in social media. The pre-processing stage focused on the removal of stop words, special characters, numbers, emoticons, HTML tags, and words with less than 4 characters.  [52] pre-processed tweets using tokenization and stop-word removal.

Event detection algorithm based on social media was proposed by [53].  The work focused on detecting foodborne disease event from Weibo (a social media network like Twitter commonly used in China). The authors adopted keyword extraction for filtering Weibo data that contained foodborne disease-related words. Support Vector Machine was also trained using ten features which include the number of followings, number of followers, user tweets, average retweets, average recommendation, average comments, average link in tweets, the length of personal description, the average length of tweets, and time of posting to filter Weibo data. The problem here is that the same keywords can refer to different things based on the context in which they are used. Also, the authors did not handle slangs/acronyms/abbreviations.

Real-time event detection from social media streams was proposed by [54]. The pre-processing stage involves removal of tweets with spam using a manually curated list of about 350 spam phrases, removal of mentions, URLs, stop words, and username. This was then followed by tokenization. Tweets were transformed into a vector form using TF-IDF. The authors did not handle slangs/acronyms/abbreviations. They also used TF-IDF for vector generation which only does a pairwise comparison but does not consider the semantics of text and the order of words.

[55] worked on the identification of vulnerable communities using hate speech detection on social media. The authors performed social media data pre-processing by removing punctuation, URLs, null values, different symbols, and retweets. Tokenization and part of speech tagging were also performed. The authors did not handle slangs/abbreviations/acronyms.

Real-time health classification from social media data with SENTINEL using deep learning was proposed by [56]. The pre-processing involves removal of email address, links, mentions, punctuations, excess spaces, translation of HTML, emoji, emoticons using emoji dictionary, removing quotes (" ") from quoted words and prefixing with a quote, splitting hashtags into individual words. However, slang/abbreviation/acronym terms were not handled.

[57] proposed corpora building automation to facilitate social media analytics. The proposed technique was tested with tweets about Indian politics and was able to achieve 85.55% in identifying the correct domain terms and entities. However, the algorithm could not recognise slangs from regional languages which underscores the significance of integrating local knowledge sources.

Social media textual content normalisation using a hybrid, and modular pipelines to normalise slangs referred to as out-of-vocabulary (OOV) to in-vocabulary (IV) was proposed by [58]. The author focused on one-to-one normalisation and one-to-many normalisation. However, the author did not focus on many-to-one normalisation. An example of many-to-one normalisation is "S.T.O.P" to "stop". This puts a limitation on the overall performance of the proposed system on the intrinsic evaluation data, besides the fact that ambiguous OOV were not addressed. [59] proposed a framework for learning a continuous representation for acronym referred to as Arc2vec using three embedding models: Average Pooling Definition Embedding (APDE), Max Pooling Definition Embedding (MPDE) and Paragraph-Like Acronym Embedding (PLAE) to resolve the ambiguity problem. The result shows that both APDE and MPDE perform better than PLAE. However, contextual information was not taken into consideration for acronym embedding.

[60] used abbreviation and lexicon to deduce demographic attributes such as gender, age, and education level. Abbreviation (or OOV) was handled with the use of a regex and recursive logic. However, their framework was unable to identify a large number of abbreviation due to the generation of contextually incorrect unrelated replacement.

[61] proposed a pre-processing framework that utilises N-gram and Hidden Markov Model for spell-checking and tweet correction. The authors tested on only 10 tweets and got an accuracy of 80%. The result cannot be generalised due to the very small number of tweets used, and the approach was not compared with different techniques.

Analysis of Twitter hashtag for users' popularity detection was proposed by [62] using Fuzzy clustering approach. While user-defined metalinguistic labels such as hashtags, comments, free-text metadata, social tagging, and geo-tagging can be useful in social media analysis, some of these labels may be ambiguous or shift in meaning over time. For instance, popular hashtags have a high tendency of being spammed. Table 1 provides an outline of the previous research efforts (that focused on handling slang or abbreviation or acronym) and the various aspects that have so far received the attention of researchers in the literature.

To the best of our knowledge, none of the previous approaches has treated all of slang, abbreviation, and acronym issues or introduced the use of localised knowledge sources for resolving ambiguity in the usage of slangs/abbreviations/acronyms originating from a specific location. Moreover, out of 15 papers (as shown in Table 1) that focused on handling slangs or abbreviations or acronyms, only 3 (20%) addressed the issue of ambiguity which is one of the main concerns in natural language processing. This limits the accuracy of the interpretation of social media feeds. For instance, a research performed by [64], used a hybrid approach with a slang classifier, where the slangs were categorised into negative or positive sentiments without contemplating the context in which the slangs have been used. The method had an overall accuracy of 90%.

Therefore, in contrast to previous efforts, this paper attempts to improve on existing approaches by finding a representation for handling noisy terms and resolving the ambiguity based on the contextual analysis of terms and their use. The proposed framework is designed to analyse the rich semantics embedded in social media streams for better understanding and interpretation to enhance the accuracy of computational techniques that rely on such analysis.

+Table 1 Overview of the characteristics of the existing research

| S/N | Sources | Handles | | | | | Includes | Research Focus |
|-----|---------|---------|---|---|---|---|----------|----------------|
| | | Slang | Abbreviation | Acronym | Disambiguation | Spell-checking | Localised knowledge source | |
| 1 | [34] | | ✓ | | | | | Sentiment analysis |
| 2 | [57] | ✓ | | | | | | Sentiment analysis/event detection |
| 3 | [60] | | ✓ | | | | | Sentiment analysis/event detection |
| 4 | [42] | ✓ | ✓ | | | | | Sentiment analysis |
| 5 | [43] | | | ✓ | ✓ | | | Sentiment analysis/event detection |
| 6 | [59] | | | ✓ | ✓ | | | Sentiment analysis/event detection |
| 7 | [6] | ✓ | | | | ✓ | | Sentiment analysis |
| 8 | [40] | ✓ | | | | | | Sentiment analysis |
| 9 | [58] | ✓ | | | | | | Sentiment analysis/event detection |
| 10 | [46] | | | ✓ | | | | Sentiment analysis |
| 11 | [65] | ✓ | | | | | | Sentiment analysis |
| 12 | [63] | ✓ | | | ✓ | ✓ | | Sentiment analysis/event detection |
| 13 | [10] | | | | | ✓ | | Sentiment analysis |
| 14 | [61] | | | | | ✓ | | Sentiment analysis |
| 15 | [44] | ✓ | | | | | | Sentiment analysis |

## 3. Methodology

This section describes the approach used to realise the Social Media Feed Pre-processing (SMFP) framework. This includes data collection, data pre-processing, and data enrichment. The architectural framework of the proposed SMFP is depicted in Figure 1.

## 3.1 Description of the SMFP framework

The proposed SMFP architectural framework consists of 3 layers, which are briefly described in the following sections.

### 3.1.1 Data layer

This layer enables the input of social media data that are scheduled for pre-processing by the SMFP framework. This is suitable for both data at rest and data in motion. Data at rest can be stored as Comma Separated Values (CSV) or JavaScript Object Notation (JSON). For data in motion, a user interface that was built around an underlying API provided by Twitter is used to collect tweets in JSON format. With JSON format, each line can be parsed as an object.

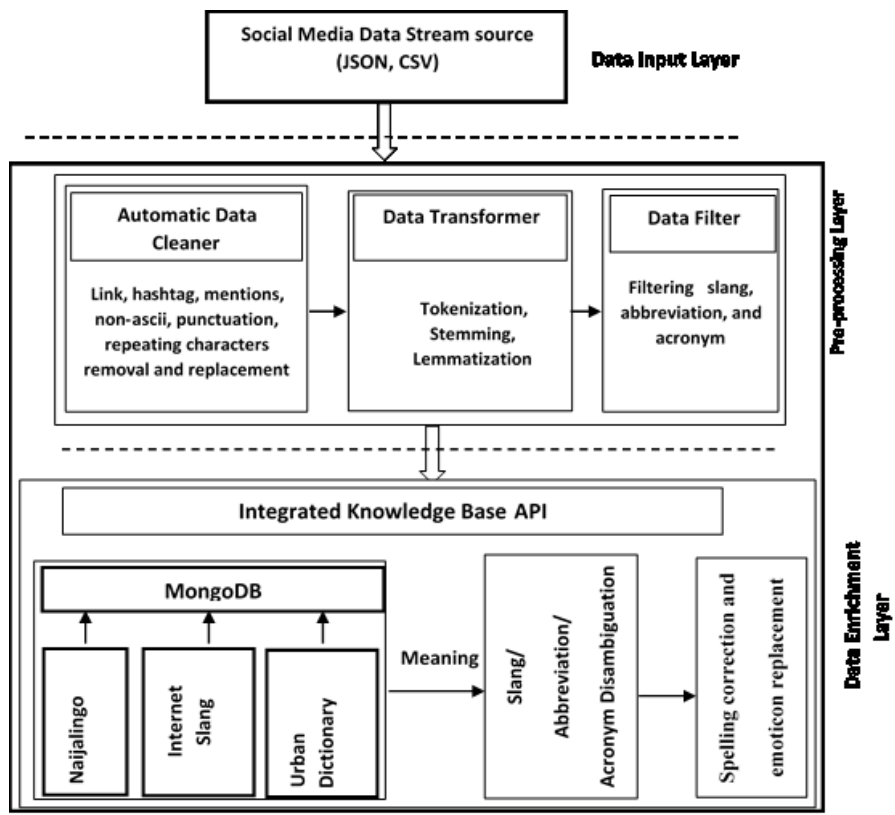

Figure 1. Proposed Social Media Feeds Pre-processing (SMFP) Framework

### 3.1.2 Data Pre-processing Layer

This layer includes three subcomponents which are data cleaner, data transformer, and data filtering. The data cleaner performs hashtag, link, non-ASCII mentions, repeating characters, punctuation removal and replacement. The data transformer is responsible for token extraction, stemming, and lemmatization. The data filter takes care of slang, abbreviation, and acronym extraction for onward transfer to the data enrichment layer for further processing.

### 3.1.3 Data Enrichment Layer

This layer houses the Integrated Knowledge Base (*ikb*) which comprises urban dictionary, Naijalingo, and internet slang. The *ikb* API allows integration of other suitable local knowledge sources for addressing slangs, abbreviations, and acronyms. The *ikb* API also caters for slang/abbreviation/acronym disambiguation, spelling correction, and emoticon replacement. The structure of a typical lexicon of the *ikb* is depicted in Figure 2.

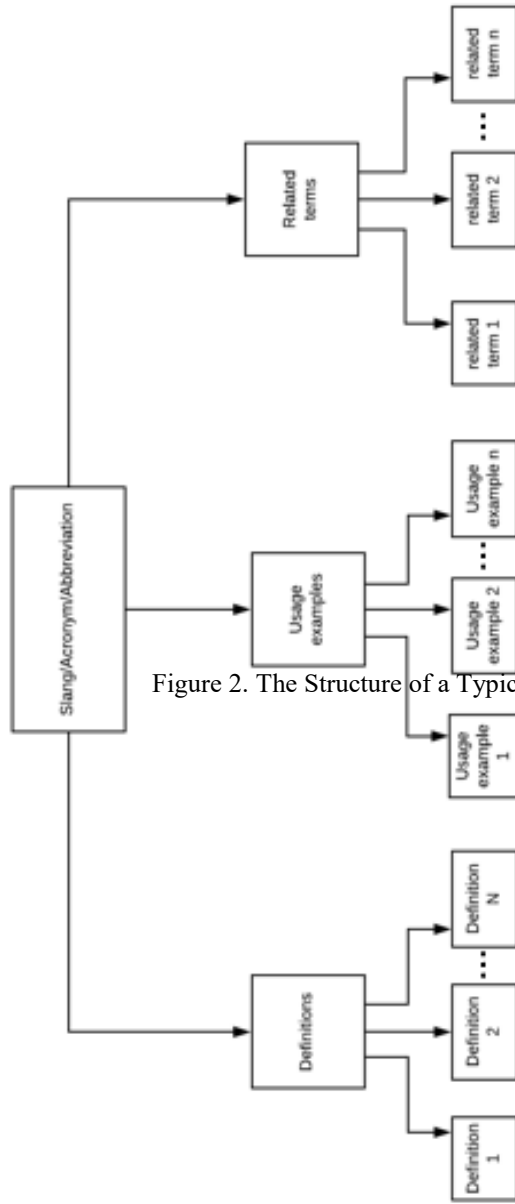

Figure 2. The Structure of a Typical Lexicon of the Integrated Knowledge Base

### 3.1.4 Process Workflow of the SMFP Framework

The operational workflow of the SMFP framework as depicted in Figure 3 is presented in this section. From the collected data stream, URLs, Tags, non-ASCII characters, and mentions are automatically removed using a regular expression. Then, the subsequent stage of the preparation will be to perform tokenization and normalization followed by slangs, abbreviations, acronyms, and emoticons filtering.

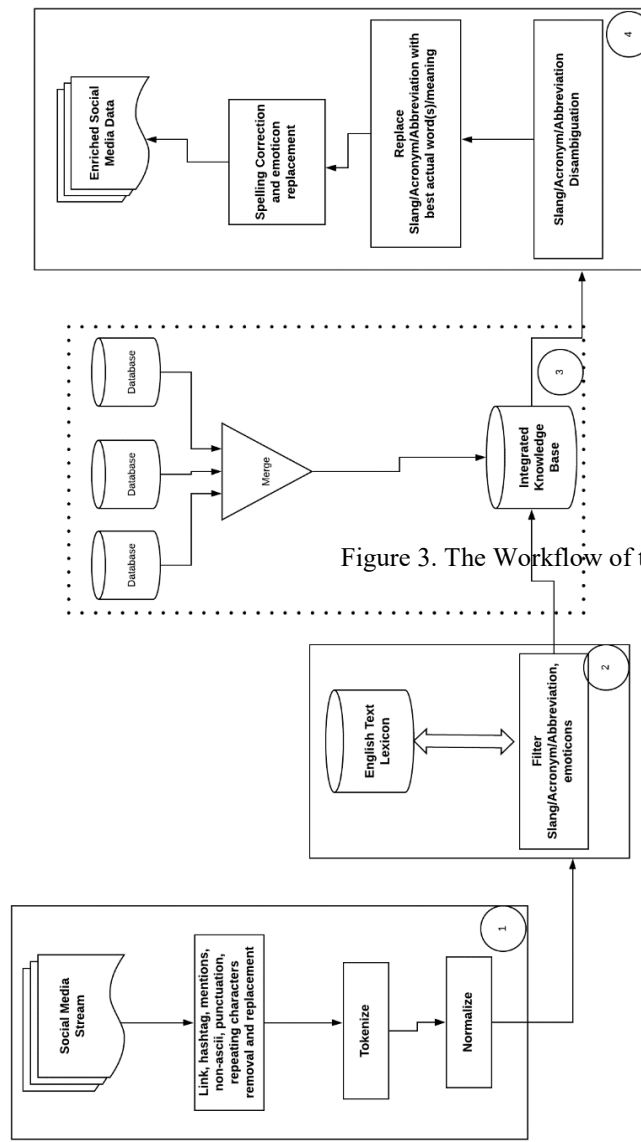

Figure 3. The Workflow of the Social Media Feed Pre-processing (SMFP) Framework

This is done with the use of corpora of English words in the NLTK library. The filtered slangs/abbreviations/acronyms are then handled by the *ikb*. Due to several meanings that may be attached to each of the slang/abbreviation/acronym terms, there is a need to disambiguate the ambiguous terms and select the best sense from the several meanings provided. This is done by adapting the Lesk algorithm to *ikb* to extract the best sense of ambiguous slangs/abbreviations/acronyms based on their context. The description of the slang/abbreviation/acronym disambiguation process is presented in Section 3.1.5. The data enrichment stage is then concluded with spellings correction and emoticon replacement. The automatic spell-checking process involves correcting typographical errors on the output from the disambiguation stage that might have escaped the *ikb* enrichment stage using the automatic spell checker library of Python. Also, the emoticon texts in the tweets are replaced with their meanings from the *ikb*.

### 3.1.5    Resolving ambiguity in slang/acronym/abbreviation

Since it is viable for each slang/acronym/abbreviation term to have more than one meaning or definition, this results in ambiguity. To resolve ambiguity in the definitions, the Lesk algorithm [66] was adapted. Lesk algorithm is a famous set of rules for word sense disambiguation. It performs disambiguation by doing a comparison between the target document and the glosses of every sense as described in WordNet. The sense with the highest cosine similarity is then treated as the best sense of the word in the target document. WordNet is a lexical database of only regular English words. WordNet does not contain slangs/acronyms/abbreviations and that necessitates the use of *ikb*. In this paper, the Lesk algorithm (see Figure 4) was adapted to determine the best sense of each slang/abbreviation/acronym observed in tweets by performing a comparison between the target slang/abbreviation/acronym with the glosses of every sense as described in *ikb*. This is illustrated with the example below.

*Example*: Consider the case of disambiguating "pale" in the tweet: **Sam, your pale has come back from work**

a. *Given the following ikb senses*

| pale[1] | Definition: The male parent that gave birth to you<br>Usage example: Ugonna your father don come from work o<br>Related term: old man, papa, patriarch, daddy, pa |
|---------|---|
| pale[2] | Definition: Having a white skin<br>Usage example: Having a pale skin isn't bad neither is having a tan skin<br>Related term: white, tan, beauty, skin, change, dull, boring |
| Pale[3] | Definition: A really bad hangover<br>Usage example: I feel so pale with what happened last night<br>Related term: whitie, pale, white, white boy |
| Pale[4] | Definition: A platonic soulmate<br>Usage example: I am really pale for Gamzee. Help<br>Related term: homestuck, quadrants |

b. *Choose the sense with the most word overlap between usage examples (usage_senses) and context* (that is, tweet in question, *sjk*)

The tweet in question, *sjk*, "Sam, your pale don come back from work" is compared with all the available usage examples (*usage_senses*) in the *ikb* relating to "pale". The score for each comparison (relatedness($st_i$,sjk)) is stored in an array (*score*). The comparison with the highest score is taken as the best score.

sam, your [pale] has come back from work

| pale[1] | Definition: The male parent that gave birth to you<br>Usage example: Ugonna your father don come from work o<br>Related term: old man, papa, patriarch, daddy, pa | 4 overlaps |
|---------|---|---|
| pale[2] | Definition: Having a white skin<br>Usage example: Having a pale skin isn't bad neither is having a tan skin<br>Related term: white, tan, beauty, skin, change, dull, boring | 1 overlap |
| pale[3] | Definition: A really bad hangover<br>Usage example: I feel so pale with what happened last night<br>Related term: whitie, pale, white, white boy | 1 overlap |
| pale[4] | Definition: A platonic soulmate<br>Usage example: I am really pale for Gamzee. Help<br>Related term: homestuck, quadrants | 1 overlap |

Usage example 1 with 4 overlaps is chosen as the most appropriate $st_i$.

c. *Map the best usage example to the corresponding definition*

The $st_i$ with the highest score is then mapped with the corresponding definition. The usage example 1 is mapped with the definition of pale[1], therefore the best sense for "pale" in this context is "the male parent that gave back to you". The pseudocode of the Adapted Lesk Algorithm is presented in Figure 4.

---

Listing 1. Adapted Lesk Pseudocode

---

**Input**: tweet text, sabt
**Output**: enriched tweet text
//Procedure to disambiguate ambiguous slang/acronyms/abbreviation in tweets by //adapting the Lesk algorithm over the usage examples of //slangs/acronym/abbreviation found in the integrated knowledge base (*ikb*)
**Notations:**
*sjk*: the current tweet being processed
slngs: slangs; acrs: acronyms; abbrs: abbreviation
sabt: a collection of slang/acronym/abbreviation terms
$w_i$: an individual slang/abbreviation/acronym in sabt
$st_i$: i[th] usage example of $w_i$ in sabt found in *ikb*

**procedure** disambiguate_all_slngs/acrs/abbrs
    **for each** $w_i$ in sabt **do**
        best_sense=disambiguate_slng/acr/abbr (w$_i$, sjk)
        display best_sense
    **end for**
**end procedure**

**function** disambiguate_slng/acr/abbr(w$_i$, sjk)
    usage_senses = extract_usage_examples(w$_i$, *ikb*)
    //usage_senses: a collection of usage examples of $w_i$ in sabt found in the *ikb*
    //usage_senses $\rightarrow$ {st$_1$, st$_2$, …st$_n$ | m $\geqq$ 1}
    int[] score // an array of semantic relatedness scores

    **for each** $st_i \in$ usage_senses of $w_i$ **do**
        **for** i= 1 to n **do**
            // n is the total number of usage examples for w$_i$
            score [i] = relatedness(st$_i$,sjk)
        **end for**
        best_score = max(score[i])
    **end for**
    $st_i$ ← best_score
    return $st_i$
    map $st_i$ with *def$_i$* //(where *def$_i$* $\in$ definition)
    replace $w_i$ in *sabt* in the tweet with *def$_i$*
**end function**

Figure 4. Adapted Lesk Pseudocode

## 4. Evaluation Experiment

The proposed SMFP framework was evaluated by benchmarking it with the General Social Media Feed Pre-processing Method (GSMFPM). The details of the proposed SMFP framework has been presented in Figure 1. Figure 5 presents the general pre-processing framework. It entails data collection, data cleaning, data transformation, and slang classification. The first three modules of the GSMFPM work like the proposed SMFP framework. The slang classifier module uses a slang dictionary to classify slangs into negative and positive sentiments. The proposed SMFP framework is different from the general pre-processing framework by leveraging *ikb* and the adapted Lesk algorithm to better interpret, analyse, and take care of ambiguity in slangs/abbreviations/acronyms found in tweets. The difference between GSMFPM and SMFP is highlighted in Table 2.

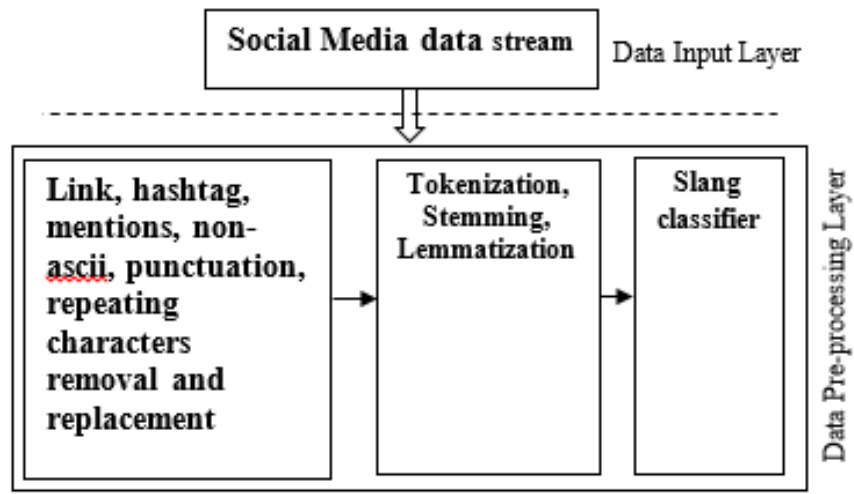

Figure 5. An Overview of the General Pre-processing Method (GSMFPM)

Table 2 Difference between GSMFPM and SMFP

| Feature | GSMFPM | SMFP |
|---|---|---|
| Abbreviation handling | No | Yes |
| Acronym handling | No | Yes |
| Disambiguation | No | Yes |
| Inclusion of Localised Knowledge Source | No | Yes |
| Spell-checking module | No | Yes |

## 4.1 Dataset description

Two datasets were used for evaluation of the proposed SMFP. The first dataset was extracted from Twitter sentiment analysis training corpus dataset[2]. The dataset originated from both Twitter sentiment corpus by Niek Sanders and the University of Michigan Sentiment Analysis on Kaggle. The dataset has 1,578,627 classified tweets with each row being marked as 1 for positive and 0 for negative. When the dataset was downloaded there was a total of 1,048,575 out of which 10% was selected by selecting every tenth record to have a total of 104,857 tweets. The selected dataset was further subdivided into training and testing datasets. Every fifth record of the 104,857 was selected as test data (i.e. 20% of the selected dataset, so that 20,971 tweets was used as test data) while the remaining was for training resulting in a total of 83,886 tweets.

The second dataset, which is called *Naija-tweets* in the previous paper [16] was extracted from tweets of Nigerian origin. The dataset focused on politics in Nigeria and contains a total of 10,000 tweets. More information on the dataset is provided in [16]. All tweets extracted had been manually classified into positive sentiment (1) or negative sentiment (0) by three computer scientists who are knowledgeable in sentiment evaluation. The sentiment results from the three experts were placed side by side and where at least two of the three result matches for a single row of a tweet, such result is selected as the sentiment for that particular tweet. 80% of the *Naija-tweets* dataset was used as training data while 20% was used as test data.

The amount of tweets extracted from Twitter sentiment analysis training corpus is much more than (approximately 10 times) that of localised dataset (*Naija-tweets*). While the Twitter sentiment analysis training corpus can be considered as a generalised/standardised dataset, the *Naija-tweets* dataset is from a specific location, Nigeria. The Twitter sentiment analysis training corpus dataset is relatively balanced because of the 104,856 tweets from Twitter sentiment analysis training corpus dataset, 52.7% was classified as positive and 47.3% as negative. The *Naija-tweets* dataset with 10,000 tweets has 64.6% classified as negative sentiments and 35.4% classified as positive sentiments, which makes it somewhat imbalanced. *Naija-tweets* dataset is about politics in Nigeria, the sentiment tilted towards more negative sentiments than positive sentiments. To correct the imbalance in *Naija-tweets* dataset in the previous paper in [16], Random Over Sampling (ROS) technique, which has better classification accuracy than both Random Under Sampling (RUS) and Synthetic Minority Oversampling Technique (SMOTE) was adopted [67]. ROS implies randomly adding instances from the minority class, in the case of *Naija-tweets* dataset, instances from positive sentiments (minority class) were randomly selected and added to make up for the imbalance. Specifically, 82.5% of the instances (totalling 2,920 tweets) from the minority class was added to make the *Naija-tweets* dataset become balanced in the ratio of 50% positive (6,460 tweets) and 50% negative (6,460 tweets).

---

[2] http://thinknook.com/Twitter-sentiment-analysis-training-corpus-dataset-2012-09-22/

## 4.2   Feature Extraction and Representation

Feature extraction has to do with the transformation of input into a set of features. For text classification, the simplest and the most commonly used features are n-grams [68-70]. In this work, two types of features namely, unigram and bi-gram were extracted from the dataset. For the unigram, a total of 76,522 unique words was extracted from the sentiment analysis training corpus dataset. After the extraction of unigrams, top K words were extracted to ward off words that occur a few times so as not to influence the classification result. A total of 50,000 unigrams were used for vector representation. There is a total of 501,026 unique bigrams in the dataset. The unique bigrams were ranked and top K amounting to 150,000 bigrams were used for vector representation.

For the *Naija-tweets* dataset, there was a total of 3,296 and 10,187 unique words and bigrams respectively. A total of 3,000 unigrams and 8,000 bigrams out of this were selected for vector representation.

## 4.3   Classifiers

For sentiment analysis, supervised or unsupervised techniques can be used to extract sentiments from tweets. With unsupervised sentiment analysis techniques, features of a given text are compared with a polarity based on discriminatory word-lexicon to classify tweets. In a supervised technique, a classifier is trained with labelled examples. In this paper, supervised techniques were used. For text classification, the three classifiers used were support vector machines (SVM), multilayer perceptron (MLP), and convolutional neural networks (CNN).

### 4.3.1   Support Vector Machines

Support Vector Machines (SVM) are discriminative classifiers that make use of statistical learning theory [71,72] to find distinct optimal separating hyperplane between two classes [73]. This is achieved by locating the maximum margin between the classes' closest points. Intuitively, an excellent separation may be accomplished by finding the hyperplane that has the biggest distance to the closest training data points of any class (so-referred to as functional margin), since in general, the higher the margin, the lower the classifier generalization error [74]. Given a set of training points $(x_i, y_i)$, where $x$ is referred to as the feature vector and $y$ is the class, the objective is to locate the maximum margin hyperplane that separates the points with $y = 1$ and $y = -1$ [75]. A discriminating hyperplane will satisfy the following inequalities:

$$w'x_i + w_0 \geq 0 \; if \; t_i = +1; \qquad (1)$$

$$w'x_i + w_0 < 0 \; if \; t_i = -1; \qquad (2)$$

where $(w', w_0)$ and $(x_i, t_i)$ represent the primal variables and training set item respectively. The distance between any point $x$ and a hyperplane is calculated by $| w'x_i + w_0 | / \| w \|$   while the distance to the origin is $|w| / \| w_0 \|$. SVM  are (1) suitable for high dimensional spaces; (2) effective when the dimensions are greater than the samples; (3) memory efficient (uses support vectors, a set of training points in the decision function); and (4) versatile, which implies that exclusive kernel features can be used for the decision function [76]. The disadvantage of the support vector machines includes (1) it results to over-fitting if the features are much larger than the samples kernel functions, to avoid this, the regularization term must be carefully chosen; (2) probability estimates are calculated using costly five-fold cross-validation since SVM do not provide probability estimates directly.

For this work, the SVM model of the Scikit-learn Machine learning library of Python was used. The regularization parameter (C), also known as the penalty parameter of the error term was set to 0.1. The SVM experiment involved the use of Unigram, Bigram, and Unigram + Bigram.

### 4.3.2 Multilayer Perceptron

A multilayer perceptron is a deep, feed-forward artificial neural network consisting of at least three layers of neurons. The multilayer perceptron is often applied to supervised learning problems. It trains on input-output pairs and learns to model the dependencies between the inputs and outputs. It learns a function $f(.): R^m \to R^o$, where $m$ represents input dimensions and $o$ represents output dimensions. Given a set of features $X = x_1, x_2, \ldots x_m$ and a target $y$, it can learn a non-linear characteristic or function approximation for both regression and classification. The input layer consists of a set of neurons $\{x_i | x_1, x_2, \ldots x_m\}$ that represent the input features. Each neuron within the hidden layer is responsible for the transformation of the values from the previous layer along with a weighted linear summation $w_1 x_1 + w_2 x_2 + \cdots w_m x_m$, this is then followed by a non-linear activation function $g(.): R \to R$ – just like the hyperbolic tangent function. The output layer receives input from the last hidden layer and transforms them into output values. The mathematical formula is as follows: Given a set of training samples $(x_1, y_1), (x_2, y_2), \ldots (x_n y_n)$, where $x_i \in R^n$ and $y_i \in \{0,1\}$, one hidden layer with one hidden neuron MLP learns the function $f(x) = W_2 g(W_1^T x + b_1) + b_2$ in which $W_1 \in R^m$ and $W_2, b_1, b_2 \in R$ are model parameters. $W_1; W_2$ constitute the weights of the input layer and hidden layer respectively; and $b_1; b_2$ represents the bias added to the hidden layer and the output layer respectively. $g(.): R \to R$ is the activation function set. This is given as:

$$g(z) = \frac{e^z - e^{-z}}{e^z + e^{-z}} \tag{3}$$

To carry out binary classification, $f(x)$ is passed through a logistic function $g(z) = 1/(1 + e^{-z})$ (where $z$ is the element of input) to get output values between $\{0, 1\}$. A threshold that is set to 0.5, implies that outputs' samples greater than or equal to 0.5 would be assigned to the positive class, and the rest to the negative class. For multiple classifications, $f(x)$ itself is a vector of (n_classes). Instead of passing via the logistic function, it passes via a softmax function [77]. Softmax is often utilised in neural networks to map un-normalised output to a probability distribution over predicted output classes. When vector elements are greater than 1 or negative and might not sum up to 1, softmax function is used for normalisation in such a way that each element $x_l$ lies within the interval {0,1} and $\sum x_i = 1$. Softmax function [78] is written as:

$$softmax(z)_i = \frac{exp(z_i)}{\sum_{l=1}^{K} exp(z_l)} \qquad (4)$$

where $z_l$ represents the i$^{th}$ input element to softmax corresponding to class $i$, and $K$ represents the number of classes. This results in a vector containing the probabilities that sample $x$ belongs to each class. The output with the highest probability is then taken as the class. MLP is efficient for complex classification by learning non-linear models. It can also learn models at real-time using partial_fit. The disadvantages include (1) MLP with hidden layers have a non-convex loss function where more than one local minimum exists. This implies that different random weight initializations lead to different validation accuracy; (2) MLP requires tuning some of the hyperparameters such as the number of hidden neurons layers, and iterations; and (3) MLP is sensitive to feature scaling.

The Keras deep learning library, running on top of TensorFlow as the back engine was used to implement the multilayer perceptron model trained with back-propagation rule and a 10-fold cross-validation procedure in this paper. The choice of Keras was informed by its ability to abstract away much of the complexity of building a neural network [79]. One-hidden layer with 500 hidden units was used. The output gives a $Pr(positive|tweet)$, which implies the probability of the tweet sentiment being positive (i.e. threshold $\geq$ 0.5). This was rounded off at the prediction step to convert to class labels 1 (positive) and 0 (negative). The model was trained using binary cross-entropy loss. The MLP experiment involved the use of Unigram, Bigram, and Unigram + Bigram

### 4.3.3 Convolutional Neural Networks

Convolutional Neural Networks (CNN) is a model of deep learning architecture that aims at learning higher-order features present in data through convolutions. In CNN, the input data is transformed through all connected layers into a set of class scores given by the output layer [80]. CNN can be used to classify a sentence into predetermined categories by considering n-grams. Given a sequence of words $w_{1:n} = w_1, w_2, \ldots, w_n$, where each word is associated with the embedding vector of dimension $d$. One-dimensional convolution with width-k is the result of moving sliding-window of size k over the sentence, and making use of the same convolution filter or kernel to every window in the sequence, i.e. a dot-product between the concatenation of the embedding vectors in a given window and a weight vector $u$, which is then followed by a non-linear activation function $g$.

Considering a window of words $w_i, \ldots, w_{i+k}$ the concatenated vector of the $i^{th}$ window is then:

$$x_i = [w_i, w_{i+1}, w_{i+k}] \in R^{k \, X \, d} \tag{5}$$

The convolution filter is applied to each window, ensuing in scalar values $r_i$, each of the $i^{th}$ window:

$$r_i = g(x_i.u) \in R \tag{6}$$

In practice, one typically applies more filters, $u_1, \ldots, u_l$, which can then be represented as a vector multiplied by a matrix U and with an addition of bias term b:

$$r_i = g(x_i.U + b) \tag{7}$$

with $r_i \in R^l, x_i \in R^{k \, X \, d}, U \in R^{k \cdot \, d \, X \, l}, and \, b \in R^l$

The Keras deep learning library, running on top of TensorFlow as the backend was used also to implement the CNN model. GloVe Twitter 27B with 200d was used for both datasets vector representation. GloVe is an unsupervised learning algorithm for obtaining word embedding or vector representation of words [81]. The idea is based on obtaining the statistics of word-word co-occurrence from a corpus, which results in representations that showcase interesting linear structures of the word vector space [82].

A vocabulary of top 180000 and 9000 words from the training dataset was used for Twitter sentiment analysis training corpus dataset and *Naija-tweets* dataset respectively. Each of the 180000 and 9000 words was represented by a 200-dimensional vector. The Embedding layer, which is the first layer is a matrix with (v + 1) x $d$ shape, where v is the vocabulary size, 180000 or 9000 as the case may be, and $d$ is the dimension of each word vector. A total of 600 filters with kernel size 3 was used for the convolution. The embedding layer was initialized with random weights from $N$ (0, 0.01), with each row representing 200-dimensional word vector for each of the words in the vocabulary. For words in the vocabulary which match the GloVe word vectors provided by the Stanford NLP group, the corresponding row of the embedded matrix was seeded from the GloVe vectors. The model was trained with binary cross-entropy loss, with 8 epochs used in running the model. Four different CNN architectures were used in the experiment, which was as follows: 1-Con-NN with a convolutional layer of 600 filters, 2-Con-NN with convolutional layers of 600 and 300 filters respectively, 3-Con-NN with convolutional layers of 600, 300, and 150 filters respectively, and 4-Con-NN was used with convolutional layers of 600, 300, 150 and 75 filters respectively.

## 5. Result and Discussion

The proposed SMFP framework was evaluated by benchmarking with the General Social Media Feed Pre-processing Method (GSMFPM) with the use of three classifiers mentioned in the previous section to extract sentiments from tweets. It has been established from the previous paper [16] that the general social media pre-processing technique does not take care of slang/abbreviation/acronym ambiguity issues. The evaluation conducted in this paper with the use of additional standardized dataset was used to complement the result in the previous paper. The result of the sentiment classification for Twitter sentiment analysis training corpus dataset is shown in Tables 3 and 4 while the analysis of an improved version of the *Naija-tweets* dataset is presented in Tables 5 and 6. The results of our previous paper [16] were based on an imbalanced *Naija-tweets* dataset but by using a balanced dataset in this paper, a better result was obtained.

Table 3. Sentiment Classification Results by SVM and MLP Based on Twitter sentiment analysis training corpus

| Method | Algorithm | Accuracy (%) Unigram | Accuracy (%) Bigram | Accuracy (%) Unigram + Bigram |
|--------|-----------|----------------------|---------------------|-------------------------------|
| GSMFPM | SVM | **78.84** | 74.16 | 76.58 |
| SMFP | | 77.61 | **74.29** | **78.90** |
| GSMFPM | MLP | 82.60 | 79.40 | 83.20 |
| SMFP | | **84.80** | **85.30** | **86.20** |

From the analysis of the result, as shown in Table 3, SVM and MLP were run on the two pre-processing methods using unigram, bigram, and the combination of unigram and bigram. The proposed SMFP framework outperformed the existing generic pre-processing method except for SVM with unigram. It should be noted that using the combination of unigram and bigram performs better than using either unigram or bigram except in the case of SVM where using the unigram with existing pre-processing method performed better. Also, using unigram alone generally performs better than using bigram alone.

Table 4. Sentiment Classification Results by CNN Based on Twitter sentiment analysis training corpus

| Method | Algorithm (Kernel size = 3) | Accuracy (%) 1-Con-NN | Accuracy (%) 2-Con-NN | Accuracy (%) 3-Con-NN | Accuracy (%) 4-Con-NN |
|--------|------------------------------|-----------------------|-----------------------|-----------------------|-----------------------|
| GSMFPM | CNN | 92.17 | 87.79 | 87.77 | 87.95 |
| SMFP | | **94.07** | **89.31** | **89.79** | **91.01** |

From Table 4, 4 different CNN architectures; 1-Con-NN, 2-Con-NN, 3-Con-NN, and 4-Con-NN, each with kernel size 3 was run on the two pre-processing methods. The proposed SMFP framework performed better than the general pre-processing method. This shows that capturing slangs/abbreviations/acronyms found in tweets and resolving ambiguity in their usage lead to improvement of subsequent analysis algorithms. Surprisingly, convolutional neural networks with one layer performed better than the two, three, and four layers for both pre-processing methods.

Table 5. Sentiment Classification Results by SVM and MLP Based on *Naija-tweets* Dataset

| Method | Algorithm | Accuracy (%) Unigram | Accuracy (%) Bigram | Accuracy (%) Unigram + Bigram |
|--------|-----------|----------------------|---------------------|-------------------------------|
| GSMFPM | SVM | **88.80** | 88.80 | 94.60 |
| SMFP | | 88.24 | **94.12** | **98.04** |
| GSMFPM | MLP | 73.64 | **88.41** | 97.46 |
| SMFP | | **73.80** | 88.20 | **99.00** |

In Table 5, an experiment from Table 3 was repeated using an improved localised dataset, tagged as *Naija-tweets*, which originated from Nigeria as described in Section 4.1, the result of the experiment revealed that the proposed SMFP framework outperformed the GSMFPM with improvement within the accuracy of the result obtained. This underscores the significance of the use of a localised knowledge base in the pre-processing of social media feeds to fully capture the noisy terms that are contained within the social media feeds originating from a specific location.

Table 6. Sentiment Classification Results by CNN Based on *Naija-tweets* Dataset

| Method | Algorithm (Kernel size = 3) | Accuracy (%) 1-Con-NN | Accuracy (%) 2-Con-NN | Accuracy (%) 3-Con-NN | Accuracy (%) 4-Con-NN |
|--------|------------------------------|------------------------|------------------------|------------------------|------------------------|
| GSMFPM | CNN | 97.81 | 97.59 | 96.30 | 95.66 |
| SMFP | | **99.78** | **99.34** | **98.47** | **96.28** |

The result presented in Table 6 is also a proof of the inclusion of localised knowledge source in the pre-processing of social media feeds from a specific origin to understand and better interpret slangs/abbreviations/acronyms stemming from such social media feeds.

## 6. Conclusion and Further Work

This paper presents an approach that enables better understanding, interpretation, and handling of ambiguity that is associated with the usage of slangs/acronyms/abbreviations in social media feeds. This is critical for subsequent learning algorithms to perform effectively and efficiently. It is an extension of the concept reported in [16] but strengthened with the use of more elaborate data in the form of a bigger, an open, and standard dataset - Twitter sentiment analysis corpus and improved version of *Naija-tweets* dataset to generate a more robust conclusion.

Also, the use of local knowledge base coupled with other knowledge sources for pre-processing of data gives a better interpretation of noisy terms such as slangs, abbreviations, and acronyms in social media feed. Thus, leading to an improvement in the accuracy of algorithms building on them. This suggests that the context from which social media feeds emanate have impact on the interpretation of terms that are contained in social media feeds, which determines the overall accuracy of operations such as sentiment analysis and event detection. The specific contributions of the paper are the following:

    a)   *An approach for improved semantic analysis and interpretation of noisy terms in social media streams*. Currently, resolving slang/abbreviation/acronym terms found in social media streams is done by getting a lexicon and plugging in the meaning without considering the context of usage of slang/abbreviation/acronym terms. The SMFP proposed in this paper leverages an integrated knowledge base (*ikb*) coupled with localised knowledge sources to capture the rich but hidden knowledge in slangs/abbreviations/acronyms, to resolve ambiguity in the usage of these noisy terms. This will foster a better understanding and interpretation of these noisy terms, which lead to an improvement in the performance of algorithms that rely on such analysis for their results.

    b)   *Reusable knowledge infrastructure for analysis of social media terms with contextually rooted meanings.* While the focus of this paper is on improving sentiment analysis in social media streams, the enrichment component of the SMFP framework can also be used for detecting events from social media streams. In addition, Integrated Knowledge Base (*ikb*) API that was created can be used as a plug-in tool by other applications for interpreting and disambiguating slang/acronym/abbreviation terms. Furthermore, the *ikb* API in

the SMFP framework allows the integration of any other local knowledge base or more knowledge sources that may suit other contexts to capture slangs/abbreviations/acronyms that have locally defined meaning.

To address the limitation of this paper, further work must focus on a better way of social media stream representation or embedding. The use of word vector for representing tweet in text classification has a problem of word order preservation and may not accurately capture the semantic context. As a result, more research effort should be directed to phrase or sentence representation for an improved result. Moreover, while Twitter is very prominent as a research data source, exploring other sources or harmonizing Twitter with other social media sources can lead to a more significant result. The concept of harmonizing multiple social media sources for sentiments analysis and event detection is still an open area of research that is worthy of further exploration since only a few approaches have considered this so far.

## Acknowledgements

The research was supported by Covenant University Centre for Research, Innovation, and Discovery; Cape Peninsula University of Technology, South Africa; The World Academy of Sciences for Research and Advanced Training Fellowship, FR Number: 3240301383; Federal University Lokoja, Nigeria; Indian Institute of Technology, Delhi.

## Conflict of Interest

The authors hereby declare that there is no conflict of interest.

Findings in Intelligent Computing Techniques. *Advances in Intelligent Computing, 79.* Singapore: Springer. doi: 10.1007/978-981-10-8633-5_9